\documentclass{SCIS2026}

\RequirePackage{xspace}

\makeatletter
\DeclareRobustCommand\onedot{\futurelet\@let@token\@onedot}
\def\@onedot{\ifx\@let@token.\else.\null\fi\xspace}

\def\ie{\emph{i.e}\onedot}

\makeatother

\definecolor{Blue}{RGB}{0,176,240}
\definecolor{Green}{RGB}{0,176,80}

\usepackage[capitalize]{cleveref}
\crefname{section}{Sec.}{Secs.}
\Crefname{section}{Section}{Sections}
\Crefname{table}{Table}{Tables}
\crefname{table}{Tab.}{Tabs.}
\Crefname{equation}{Equation}{Equations}
\crefname{equation}{Eqn.}{Eqns.}

\usepackage{lipsum}
\usepackage{bm}
\usepackage{caption}
\usepackage{tabularx}

\usepackage[table]{xcolor}
\usepackage{amsmath}
\usepackage[abs]{overpic}
\newcommand{\tabincell}[2]{\begin{tabular}{@{}#1@{}}#2\end{tabular}}

\begin{document}
\ArticleType{RESEARCH PAPER}
\Year{2025}
\Month{January}
\Vol{68}
\No{1}
\DOI{}
\ArtNo{}
\ReceiveDate{}
\ReviseDate{}
\AcceptDate{}
\OnlineDate{}
\AuthorMark{}
\AuthorCitation{}

\title{4DHumanDiff: Direct Text-to-4DGS Generation for Consistent 360-Degree Dynamic Humans}{Renlong WU, Haoran CHEN, Yuxiang WEI, Xiaowei JIN, Wangmeng ZUO, and Hui LI. 4DHumanDiff: Direct Text-to-4DGS Generation for Consistent 360-Degree Dynamic Humans}

\author{Renlong Wu}{}
\author{Haoran Chen}{}
\author{Yuxiang Wei}{}
\author{Xiaowei Jin}{xiaowei.jin@hit.edu.cn}
\author{Wangmeng Zuo}{}

\author{Hui Li}{}

\address{School of Computer Science and Technology, Harbin Institute of Technology, Harbin 150001, China}

\abstract{
Generating high-quality 360-degree dynamic human assets from text prompts is challenging. Existing methods usually synthesize monocular or multi-view videos first and then fit a 4D representation, which is expensive and often causes incomplete geometry or view-inconsistent renderings.   We present 4DHumanDiff, a diffusion framework that directly generates dynamic humans represented by 4D Gaussian Splatting (4DGS) from text prompts. By modeling the structured 4D representation space end-to-end, 4DHumanDiff avoids video pre-generation and per-scene reconstruction, making it better suited for view-consistent and temporally coherent asset generation. The model uses a 3D U-Net backbone with temporal attention for motion-aware generation. We further construct a large-scale text-to-4DGS dataset with 60,000 high-quality pairs, and introduce 2D regularization and training-free 4D interpolation to improve rendering quality and motion smoothness. Experiments show that 4DHumanDiff generates consistent 360-degree dynamic humans within one minute, achieves better temporal and multi-view consistency, and reduces inference time by more than 10$\times$.
}


\maketitle

\section{Introduction}

High-quality 360-degree dynamic human assets are important for applications such as virtual try-on~\cite{ju2023humansd,santesteban2021self,santesteban2022ulnef,wang2025mv} and immersive telepresence~\cite{ju2023direct,ju2023human,liu2022audio,zhu2023taming}. 
However, creating such assets with conventional pipelines is often time-consuming, as it usually requires careful modeling, rigging, and manual animation~\cite{ren2024l4gm}. 
With the rapid progress of generative models~\cite{huang2025voyager,zhang2023adding,yang2024cogvideox}, generating dynamic humans directly from text prompts has become increasingly attractive. 
Still, most existing generative approaches mainly aim at producing images or videos, rather than 360-degree dynamic human assets that can be rendered consistently from arbitrary viewpoints.

Some methods~\cite{wan2025wan,zhao2024genxd,yao2025sv4d,zhang20244diffusion,ren2024l4gm} address this problem with a two-stage pipeline. 
They first synthesize monocular or multi-view videos from text, and then reconstruct a 4D representation for each scene through a separate fitting process~\cite{hu2024gauhuman,zhan2025real,qian20243dgs,li2024animatable,zhan2025r3,ren2024l4gm}. 
Although effective to some extent, this design has several limitations. 
First, both video synthesis and per-scene 4D fitting are computationally expensive, leading to high overall cost. 
Second, when only monocular videos are generated, the limited viewpoint coverage makes it difficult to recover complete 360-degree geometry. 
Third, even when multi-view videos are synthesized, inconsistencies across viewpoints often remain, which further degrade the quality of the reconstructed 4D assets. 
These limitations suggest that generating 360-degree dynamic humans through an intermediate video representation is not an ideal solution.

\begin{figure*}[t!]
    \centering
    \includegraphics[width=0.99\linewidth]{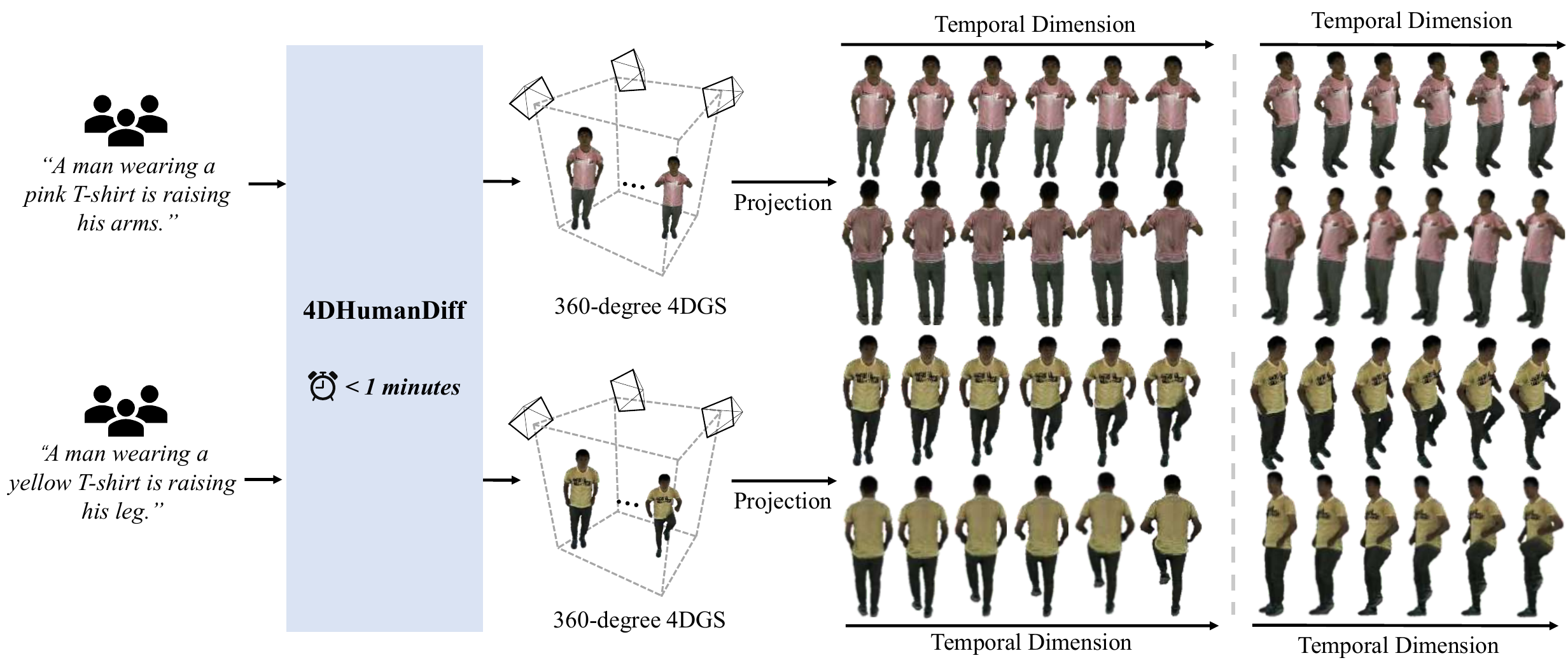}
    \vspace{-2mm}
    \caption{4DHumanDiff directly generates 4DGS-based 360-degree dynamic humans within one minute from user-provided text prompts.}
    \label{fig:intro_overview}
\end{figure*}

In this work, we instead formulate dynamic human generation directly in the 4D representation space. 
Rather than generating videos first and reconstructing later, we directly generate dynamic humans represented by 4D Gaussian Splatting (4DGS)~\cite{wu20244d,yang2023real} from text prompts, as shown in \cref{fig:intro_overview}.
This formulation is particularly suitable for human generation. 
Compared with general dynamic scenes, the human body has a clearer underlying structure, \ie, different body parts form an organized whole, and their motions are also strongly correlated over time. 
Such spatial organization and temporal regularity are difficult to preserve when generation is performed only in the video space, but can be modeled more naturally in a structured 4D representation. 
Meanwhile, directly using 4DGS as the generation target better aligns the generation process with the final goal of 360-degree dynamic human asset creation. 
It removes the need for video pre-generation and per-scene 4D fitting, reduces the overall computational burden, and provides a more suitable basis for generating view-consistent and temporally coherent results.

Based on this idea, we propose 4DHumanDiff, a diffusion framework for direct text-to-4DGS human generation, as shown in \cref{fig:pipeline}. 
Our model is built on a 3D diffusion backbone~\cite{nichol2021improved,dhariwal2021diffusion,zhang2024gaussiancube} and incorporates temporal self-attention to model motion consistency over time. 
A key part of this work is the construction of a large-scale text-to-4DGS dataset for dynamic humans. 
Built upon MVHumanNet~\cite{xiong2024mvhumannet}, the dataset contains 60,000 high-quality text--4DGS pairs and enables training 4DHumanDiff from scratch. 
In addition, we introduce a 2D regularization term and a training-free 4D interpolation strategy to improve rendering quality and motion smoothness. 
With these designs, 4DHumanDiff can generate high-quality 360-degree dynamic human assets within one minute.

Experiments show that 4DHumanDiff achieves strong performance in both quality and efficiency. 
It produces more consistent results than existing methods in both temporal and view dimensions, while reducing the overall generation time by more than 10$\times$. 
These results demonstrate the potential of direct text-to-4DGS generation as an effective alternative to existing video-first pipelines for dynamic human generation.

The main contributions of this work are summarized as follows:
\begin{itemize}
    \item We present 4DHumanDiff, a diffusion framework that directly generates 4DGS-based 360-degree dynamic humans from text prompts, avoiding the video-first and reconstruction-later pipeline used in previous methods.
    
    \item We construct a large-scale text-to-4DGS dataset for dynamic humans, containing 60,000 spatially and temporally structured 4DGS human samples, which enable training the proposed 4DHumanDiff from scratch.
    
    \item We introduce simple and effective designs, including temporal self-attention, a 2D regularization term, and a training-free 4D interpolation strategy, to improve generation quality and motion smoothness. 

    \item Extensive experiments show that 4DHumanDiff produces high-quality 360-degree dynamic human generation within one minute, achieving superior temporal and multi-view consistency while reducing inference time by more than 10$\times$.
\end{itemize}

\section{Related Work}
\label{sec:related}

\subsection{3D Representation}
A 3D model can be represented in implicit or explicit manners.
Early neural radiance field (NeRF) approaches~\cite{NeRF,barron2021mip,martin2021nerf,yang2023freenerf,pumarola2021d,muller2022instant} adopt implicit formulations, encoding scenes without explicitly defined geometry. 
These methods rely on volumetric ray marching to optimize a continuous scene representation, resulting in significant computational overhead.
3D Gaussian Splatting~\cite{kerbl20233d,wu2024dual,fu2023colmap,chung2023depth,dou2024cosseggaussians,zhuang2024tip,yu2023mip,morgenstern2023compact} models world with explicit Gaussian ellipsoids, achieving impressive reconstruction fidelity and real-time rendering performance.
However, these methods typically require per-scene optimization, leading to substantial computational cost.
Recent works~\cite{zhang2024gaussiancube,lan2024ln3diff,zhou2024diffgs,roessle2024l3dg,cai2025baking,liu2024novelgs,go2025videorfsplat,tang2024lgm} explore generalizable 3D reconstruction that can recover 3D models within a few seconds.

\subsection{Video-to-4D Reconstruction}
Most 4D reconstruction methods~\cite{ fridovich2023k,guo2023forward, feng20233d, somraj2024factorized,duan20244d,lu20243d,luiten2023dynamic,yang2023real,kratimenos2023dynmf,huang2024sc,lin2024gaussian,li2024spacetime,shawswings,sun20243dgstream,4dgaussians,yang2024deformable,bae2024per,mihajlovic2024splatfields,4d-rotor20244d,chu2024dreamscene4d,katsumata2024compact} introduce an implicit or explicit deformation model for object motion representation to reconstruct 4D models from multi-view videos.
However, for single-view video inputs, the problem becomes ill-posed and some studies introduce data-driven priors, such as depth maps~\cite{lee2023fast,yang48953364d}, optical flows~\cite{tian2023mononerf,gao2024gaussianflow,guo2024motion,liu2024modgs,zhu2024motiongs,lee2023fast,wang2024gflow}, tracks~\cite{wang2024shape,seidenschwarz2024dynomo,stearns2024dynamic,lei2024mosca}, and generative models~\cite{wu2025sc4d, chu2024dreamscene4d,yu20244real,zeng2025stag4d}.
Recent generalizable feed-forward 4D reconstruction models~\cite{tang2024lgm,ren2024l4gm,lv2025physgm,ma20254d,liang2024feed,sabathier2025lim} have been proposed to avoid per-scene optimization, accelerating the reconstruction.
For human reconstruction, most works~\cite{zhan2025real,hu2024gaussianavatar,hu2024gauhuman,hu2024surmo,jena2023splatarmor,jung2023deformable,kocabas2024hugs,lei2024gart,li2023human101,li2024gaussianbody,li2024animatable,liu2024gea,liu2024animatable,moreau2024human,niu2025anicrafter,pang2024ash,qian20243dgs,zhan2025r3,zheng2024gps,zheng2024physavatar,zielonka2025drivable,masuda2024generalizable,li2024ghunerf} introduce SMPL~\cite{loper2023smpl} prior for better recovering dynamic human assets, which parameterizes the human body as individual shape components and motion-related human poses through 3D mesh scanning and PCA.
Despite the remarkable progress, they rely on user-captured high-quality videos for scene recovery, which limits their flexibility and application.

\subsection{Text-to-4D Generation}
%
Most text-to-4D methods~\cite{zhao2024genxd,yao2025sv4d,zhang20244diffusion,liu2025free4d,sun2024dimensionx,jiang2023consistent4d,zeng2024stag4d,yin20234dgen} heavily rely on pretrained video diffusion models.
Based on score distillation sampling~\cite{poole2022dreamfusion,lin2023magic3d,yu2023text,wang2023prolificdreamer}, some approaches~\cite{jiang2023consistent4d,zeng2024stag4d,yin20234dgen,ren2023dreamgaussian4d} inject generative priors from video diffusion models~\cite{singer2022make,luo2023videofusion,wan2025wan,hong2022cogvideo,yang2024cogvideox,shi2023mvdream,blattmann2023stable} to synthesize 4D assets through per-object optimization.
Alternatively, multi-view video-based methods~\cite{zhao2024genxd,yao2025sv4d,zhang20244diffusion,liu2025free4d,sun2024dimensionx} first generate multi-view videos from text prompts and then perform 4D reconstruction.
However, these methods generally perform unsatisfactorily when generating 360-degree dynamic human assets, due to insufficient multi-view consistency and high computation cost from video diffusion models or per-scene 4D optimization.
Our 4DHumanDiff exploits the diffusion model to directly generate 4DGS-based 360-degree dynamic human assets from user-provided text prompts, without the necessity for any pretrained diffusion model priors.

\begin{figure*}[t!]
    \centering
    \includegraphics[width=0.99\linewidth]{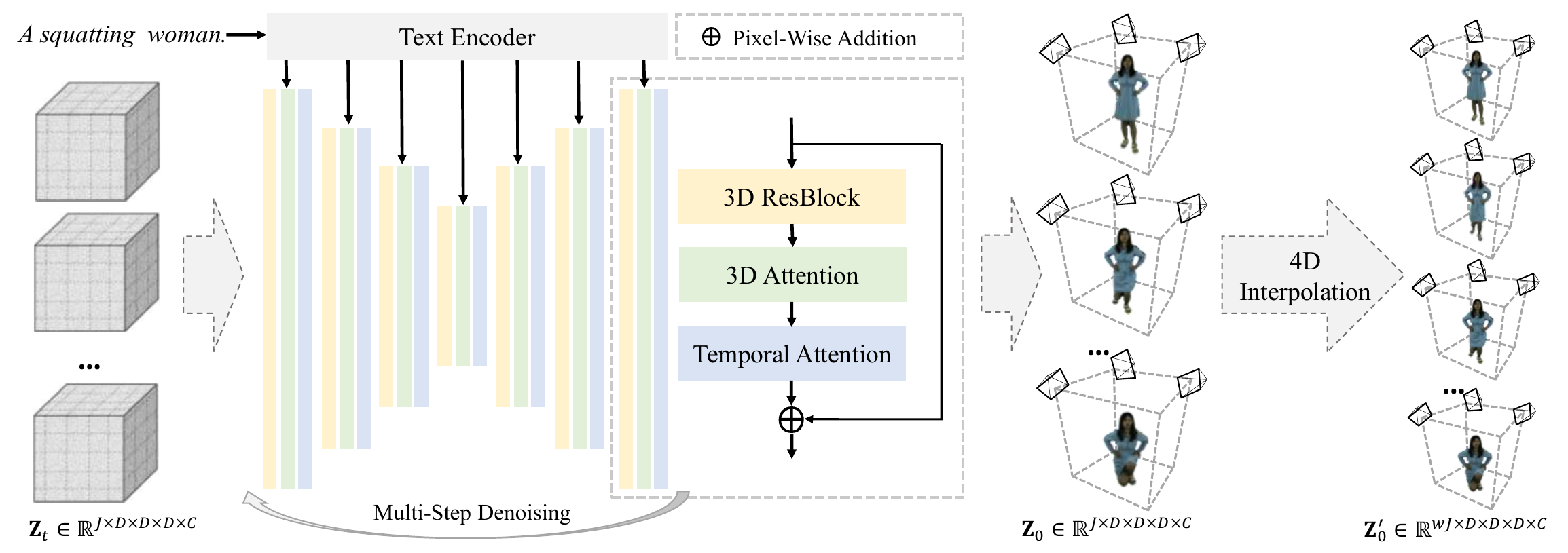}
    \vspace{-2mm}
    \caption{Overview of the proposed 4DHumanDiff. It takes user-provided text prompts as input and directly generates 4DGS-based 360-degree dynamic human assets. The 4D U-Net is built upon a 3D U-Net, where temporal attention layers are incorporated for better temporal consistency. Besides, 4D interpolation is applied to enhance motion smoothness.}
    \label{fig:pipeline}
\end{figure*}

\section{Proposed Method}
\label{sec:method}

\subsection{Preliminaries: 4D Gaussian Splatting}
Given a set of images with camera poses, 3D Gaussian Splatting (3DGS)~\cite{kerbl20233d} represents a scene as a collection of 3D Gaussian primitives, where an adaptive density control strategy is used to regulate the number of Gaussians. 
Each 3D Gaussian is parameterized by $\{\mathbf{x}, \mathbf{r}, \mathbf{s}, \alpha, \mathbf{c}\}$, where $\mathbf{x} \in \mathbb{R}^{3}$ denotes its center in world space, $\mathbf{r} \in \mathbb{R}^{4}$ and $\mathbf{s} \in \mathbb{R}^{3}$ define its rotation and scale, $\alpha \in \mathbb{R}$ denotes opacity, and $\mathbf{c} \in \mathbb{R}^{3}$ represents the spherical harmonics coefficients for color.
To model object motion in dynamic scenes, 4D Gaussian Splatting (4DGS)~\cite{wu20244d,yang2023real} defines a set of canonical Gaussians at a reference timestamp and deforms them to other timestamps. 
Let $\mathbf{C}$ and $\mathbf{G}_{j}$ denote the canonical Gaussians and the dynamic Gaussians at the $j$-th timestamp, respectively. 
The deformation can be written as,
\begin{equation}
\mathbf{G}_{j} = \mathcal{F}(\mathbf{C}, j; \Theta_{\mathcal{F}}).
\end{equation}
Here, $\mathcal{F}$ denotes the deformation function parameterized by $\Theta_{\mathcal{F}}$.
Considering the structural characteristics of the human body, human 4D reconstruction methods~\cite{hu2024gauhuman,zhan2025real,qian20243dgs,li2024animatable,zhan2025r3} typically incorporate SMPL priors~\cite{loper2023smpl} into the representation or optimization process.

Collectively, a 4DGS $\mathbf{G} \in \mathbb{R}^{J \times N \times C}$ can be regarded as a sequence of 3DGSs, \ie, $\mathbf{G} = \{\mathbf{G}_{j}\}_{j=1}^{J}$, where $J$ is the number of timestamps, $N$ is the number of Gaussians, and $C$ is the number of feature channels.

\subsection{Text-to-4DGS Human Dataset}

To train 4DHumanDiff, we construct a text-to-4DGS dataset for 360-degree dynamic human generation based on publicly available multi-view dynamic human scans~\cite{xiong2024mvhumannet,li2025mvhumannet++,wang20244d,icsik2023humanrf,cheng2023dna,cheng2022generalizable,yu2020humbi}. 
Considering the number of scans, camera views, and motion diversity, we choose MVHumanNet~\cite{xiong2024mvhumannet} as the primary data source. 
For each human scan with 48 camera views, we first use Gemini 1.5~\cite{team2024gemini} to select clips with notable motion, and then generate paired text descriptions and 4DGS representations for each clip.

For text annotation, we feed the front-view video of each clip into a vision-language model (\ie, Gemini 1.5~\cite{team2024gemini}) and ask it to describe the human subject. 
The prompt is: ``Please provide a detailed description of the person's appearance and body motions in this video, including clothing, pose, and action dynamics over time. Focus on the human subject and avoid unnecessary background details.''
A straightforward way to obtain the corresponding 4DGS data is to fit each clip with existing 4D human reconstruction methods~\cite{hu2024gauhuman,zhan2025real,qian20243dgs,li2024animatable,zhan2025r3}, such as GauHuman~\cite{hu2024gauhuman}. 
However, directly using these reconstructed results as diffusion targets is not ideal for three reasons. 
First, the number of Gaussians varies across clips due to adaptive density control. 
Second, the global scales of reconstructed 4DGSs are inconsistent. 
Third, the Gaussians lack a predefined spatial ordering, leading to a disorganized spatial layout~\cite{zhang2024gaussiancube}. 
These issues make it difficult to build a unified and structured representation for text-to-4DGS generation.

To address them, we introduce three processing steps. 
First, we replace adaptive density control with the densification-constrained fitting strategy~\cite{zhang2024gaussiancube}, so that all clips share a fixed number of Gaussians while maintaining fitting quality, \ie, $\mathbf{G}\in \mathbb{R}^{J \times N_{\max} \times C}$, where $N_{\max}$ is a predefined hyperparameter. 

Second, we normalize $\mathbf{G}$ to a unit scale by jointly rescaling Gaussian centers and scales, yielding the normalized 4DGS $\hat{\mathbf{G}}$. 
Let $\{\mathbf{x}_{j,n}\mid \mathbf{x}_{j,n}=(x_{j,n},y_{j,n},z_{j,n}),\, j\in[1,J],\, n\in[1,N]\}$ and $\{\mathbf{s}_{j,n}\mid j\in[1,J],\, n\in[1,N]\}$ denote the Gaussian centers and scales of $\mathbf{G}$, respectively. 
The normalized center $\hat{\mathbf{x}}_{j,n}$ and scale $\hat{\mathbf{s}}_{j,n}$ of $\hat{\mathbf{G}}$ are computed as,
\begin{equation}
\left\{
\begin{aligned}
\mathbf{x}_{j,n}' &= \mathbf{x}_{j,n} - \frac{1}{J}\frac{1}{N} \sum_{j=1}^{J} \sum_{n=1}^{N} \mathbf{x}_{j,n}, \\
d_{\max} &= \max_{j,n}\big(\max(|x'_{j,n}|, |y'_{j,n}|, |z'_{j,n}|)\big), \\
\hat{\mathbf{x}}_{j,n} &= \frac{1}{2d_{\max}} \, \mathbf{x}_{j,n}',\\
\hat{\mathbf{s}}_{j,n} &= \frac{1}{2d_{\max}} \, \mathbf{s}_{j,n}.
\end{aligned}
\right.
\end{equation}
This normalization maps all Gaussian centers into a unit cube and reduces scale variation across samples.

Third, we reorganize $\hat{\mathbf{G}}$ into an explicit structured representation $\mathbf{G}'$ that preserves both spatial layout and temporal correspondence. 
Specifically, we first arrange the Gaussians at the first timestamp, $\hat{\mathbf{G}}_{1}$, into a predefined voxel grid via Optimal Transport~\cite{zhang2024gaussiancube}, thereby establishing a fixed spatial index for all Gaussians. 
Importantly, this voxelization is performed only once at the first timestamp, rather than independently for each timestamp. 
For the remaining timestamps, we do not re-voxelize the Gaussians. 
Instead, each Gaussian at timestamp $j$ is assigned to the same grid position as its corresponding Gaussian at the first timestamp according to the temporal binding relationship. 
In this way, all timestamps share the same structural layout, while the Gaussian features at each grid location evolve over time to encode motion. 
This design explicitly preserves temporal correspondence.

After these steps, the resulting structured 4DGS is represented as $\mathbf{G}' \in \mathbb{R}^{J\times D \times D \times D \times C}$, where $D=\sqrt[3]{N_{\max}}$. 
This structured representation provides a unified space for diffusion modeling and allows us to directly adopt a standard U-Net backbone without complicated architectural modifications. 
For simplicity, we denote the final structured 4DGS by $\mathbf{G}$.

\subsection{Text-to-4DGS Diffusion Model}

Our text-to-4DGS diffusion model aims to learn the distribution $p(\mathbf{G})$ of the structured 4DGS representation $\mathbf{G}$. 
The generation process is formulated as the reverse of a discrete-time Markov diffusion process. 
In the forward process, Gaussian noise is progressively added to a clean sample $\mathbf{G}^0 \sim p(\mathbf{G})$, yielding a sequence of noisy samples $\{\mathbf{G}^t \mid t \in [0,T]\}$. It can be written as,
\begin{equation}
\mathbf{G}^t = \alpha_t \mathbf{G}^0 + \sigma_t \boldsymbol{\epsilon},
\end{equation}
where $\boldsymbol{\epsilon} \sim \mathcal{N}(0, \mathbf{I})$ denotes Gaussian noise, and $\alpha_t$ and $\sigma_t$ are the noise schedule coefficients. 
After sufficient diffusion steps, $\mathbf{G}^T$ approaches a standard Gaussian distribution. 
The generation process reverses this diffusion trajectory by iteratively denoising from $\mathbf{G}^T \sim \mathcal{N}(0, \mathbf{I})$ to $\mathbf{G}^0$.
Given an intermediate sample $\mathbf{G}^t$ and the text feature $\mathbf{Y}_{text} \in \mathbb{R}^{L \times E}$, the diffusion network $\mathcal{D}$ predicts the clean representation $\hat{\mathbf{G}}^0$.
It can be written as,
\begin{equation}
\hat{\mathbf{G}}^0 = \mathcal{D}(\mathbf{G}^t, \mathbf{Y}_{text}; \Theta_{\mathcal{D}}),
\end{equation}
where $\Theta_{\mathcal{D}}$ denotes the parameters of $\mathcal{D}$.
$L$ and $E$ are the length and channel dimension of $\mathbf{Y}_{text}$, respectively.

\noindent\textbf{Model Architecture.}
Benefiting from the structured organization of $\mathbf{G}$, we adopt 3D convolutions~\cite{ji20123d,wu2019pointconv} to model spatial structure and temporal attention~\cite{blattmann2023stable,ren2024l4gm} to capture motion dependency across timestamps, respectively.
Specifically, following GaussianCube~\cite{zhang2024gaussiancube}, we extend the asymmetric 2D U-Net~\cite{nichol2021improved,dhariwal2021diffusion} into a 3D U-Net by replacing the original 2D convolution, attention, upsampling, and downsampling operations with their 3D counterparts. 
On top of this backbone, we insert temporal attention layers after each block to explicitly model temporal coherence, resulting in our 4D U-Net, as shown in \cref{fig:pipeline}. 
Furthermore, under the widely adopted assumption in 4D reconstruction~\cite{duan20244d,lu20243d,luiten2023dynamic,4dgaussians,yang2024deformable,wang2025shape,wu2024deblur4dgs} that only Gaussian positions and shapes vary over time, we keep the color and opacity attributes temporally shared to further improve temporal consistency.

\noindent\textbf{Model Optimization.}
Our model is optimized with a two-stage training strategy. 
We first train the 3D U-Net backbone to capture the spatial structure of the 4DGS representation, and then jointly optimize the whole network together with the temporal attention layers to model temporal coherence. 
This strategy stabilizes training and improves text-consistent 4D generation.
Our training objective consists of a 4D reconstruction loss $\mathcal{L}_{4d}$ and a 2D regularization term $\mathcal{L}_{2d}$, which can be written as, 
\begin{equation}
\label{eqn:loss_term}
\mathcal{L} = \mathcal{L}_{4d} + \lambda \mathcal{L}_{2d}.
\end{equation}
$\lambda$ is the regularization weight.
The 4D reconstruction loss $\mathcal{L}_{4d}$ supervises the prediction of the clean target $\mathbf{G}^0$, \ie, 
\begin{equation}
\mathcal{L}_{4d} = \left\| \mathcal{D}(\mathbf{G}^t, \mathbf{Y}_{text}; \Theta_{\mathcal{D}}) - \mathbf{G}^0 \right\|_1.
\end{equation}
To further improve the visual quality of rendered multi-view videos, we introduce a 2D regularization term $\mathcal{L}_{2d}$.
It can be written as,
\begin{equation}
\mathcal{L}_{2d} = \mathcal{L}_{1} + \mathcal{L}_{vgg}.
\end{equation}
Here, $\mathcal{L}_{1}$ measures the pixel-wise difference between the rendered $M$-view videos $\mathbf{V}_{pred}$ and the ground-truth videos $\mathbf{V}_{gt}$, which can be written as,
\begin{equation}
\mathcal{L}_{1} = \|\mathbf{V}_{pred} - \mathbf{V}_{gt}\|_1.
\end{equation}
The perceptual loss $\mathcal{L}_{vgg}$ is defined as,
\begin{equation}
\mathcal{L}_{vgg} = \|\phi(E(\mathbf{V}_{pred})) - \phi(E(\mathbf{V}_{gt}))\|_1,
\end{equation}
where $\phi$ denotes the pre-trained VGG-19~\cite{simonyan2014very} network, and $E(\cdot)$ denotes the patch selection operator that retains only patches containing human regions. 
This operation reduces background interference in perceptual supervision and provides more targeted supervision on human appearance and motion.

\noindent\textbf{4D Interpolation.}
Users generally prefer temporally smooth dynamic results. 
A straightforward solution is to apply video interpolation~\cite{RIFE,EMAVFI,UPRNet,BiFormer,wu2024dual} to rendered videos, or to train an additional interpolation model in 4D space, such as L4GM~\cite{ren2024l4gm}. 
However, the former does not explicitly guarantee multi-view consistency, while the latter introduces extra training and inference cost. 
In contrast, our structured 4DGS explicitly preserves temporal correspondence across timestamps, which allows us to directly interpolate neighboring Gaussian states for smoother motion in a training-free manner.
It can be written as,
\begin{equation}
\small
\label{eq:4d_interpolation}
\mathbf{G}_{j,i} = \left(1 - \frac{i-1}{W-1}\right)\odot \mathbf{G}_{j-1} + \frac{i-1}{W-1}\odot \mathbf{G}_{j}.
\end{equation}
Here, $\mathbf{G}_{j,i}$ denotes the $i$-th interpolated Gaussian representation between the $(j-1)$-th and $j$-th timestamps, and $W$ is the interpolation factor. 
Since the interpolation is performed on temporally aligned Gaussian representations, it naturally preserves view consistency while improving motion smoothness.

\begin{table*}
\centering
\caption{Quantitative comparison results. Existing methods first synthesize monocular or multi-view videos from text prompts, followed by per-scene 4D reconstruction. Compared with the video-based two-stage manner, our 4D-based end-to-end 4DHumanDiff performs better while being over 10$\times$ faster.$\uparrow$ denotes the higher metric the better, and $\downarrow$ denotes the lower one the better. The best metric is \textbf{boldfaced}, and the second one is \underline{underlined}.
}
\scalebox{0.80}{\begin{tabular}{c c c c c c c c c}
\toprule
\tabincell{c}{Category}
  & \tabincell{c}{Methods} & \tabincell{c}{Image \\ Quality $\uparrow$ } & \tabincell{c}{Motion \\ Smoothness $\uparrow$} & \tabincell{c}{Subject \\ Consistency $\uparrow$} & \tabincell{c}{Multi-View \\ Consistency $\uparrow$} & \tabincell{c}{Overall \\ Consistency $\uparrow$} & \tabincell{c}{Inference \\ Time (Minutes)$\downarrow$}\\
 \midrule
\multirow{5}{*}{\begin{tabular}[c]{@{}c@{}} Text-to-Video + \\ Video-to-4DGS \end{tabular}} &  Wan 2.1~\cite{wan2025wan} & 0.332 & 0.985 & 0.896 & 0.667 & 0.167 & $>$ 25\\
 \multicolumn{1}{c}{} &  L4GM~\cite{ren2024l4gm} & 0.390 & 0.994 & \underline{0.929} & 0.702 & 0.177 &  $>$ \underline{11}\\
  \multicolumn{1}{c}{} &  GenXD~\cite{zhao2024genxd}  & 0.405 & 0.978 & 0.866 & 0.798 & \underline{0.187} &  $>$ 42 \\
  \multicolumn{1}{c}{} &  SV4D 2.0~\cite{yao2025sv4d} & 0.387 & \underline{0.995} & 0.877 & \underline{0.800} & 0.155 & $>$ 45 \\
  \multicolumn{1}{c}{} &  4Diffusion~\cite{zhang20244diffusion} & \textbf{0.473} & 0.989 & 0.909 & 0.740  & 0.180 & $>$ 47 \\
  \midrule
\multirow{1}{*}{\begin{tabular}[c]{@{}c@{}} Text-to-4DGS \end{tabular}} &  4DHumanDiff (Ours) & \underline{0.445} & \textbf{0.998} & \textbf{0.984} & \textbf{0.862} &  \textbf{0.188}& $<$\textbf{1} \\
\bottomrule
\end{tabular}}
\label{tab:compared_methods}
\end{table*}

\begin{figure*}[t!]
    \centering
    \includegraphics[width=0.99\linewidth]{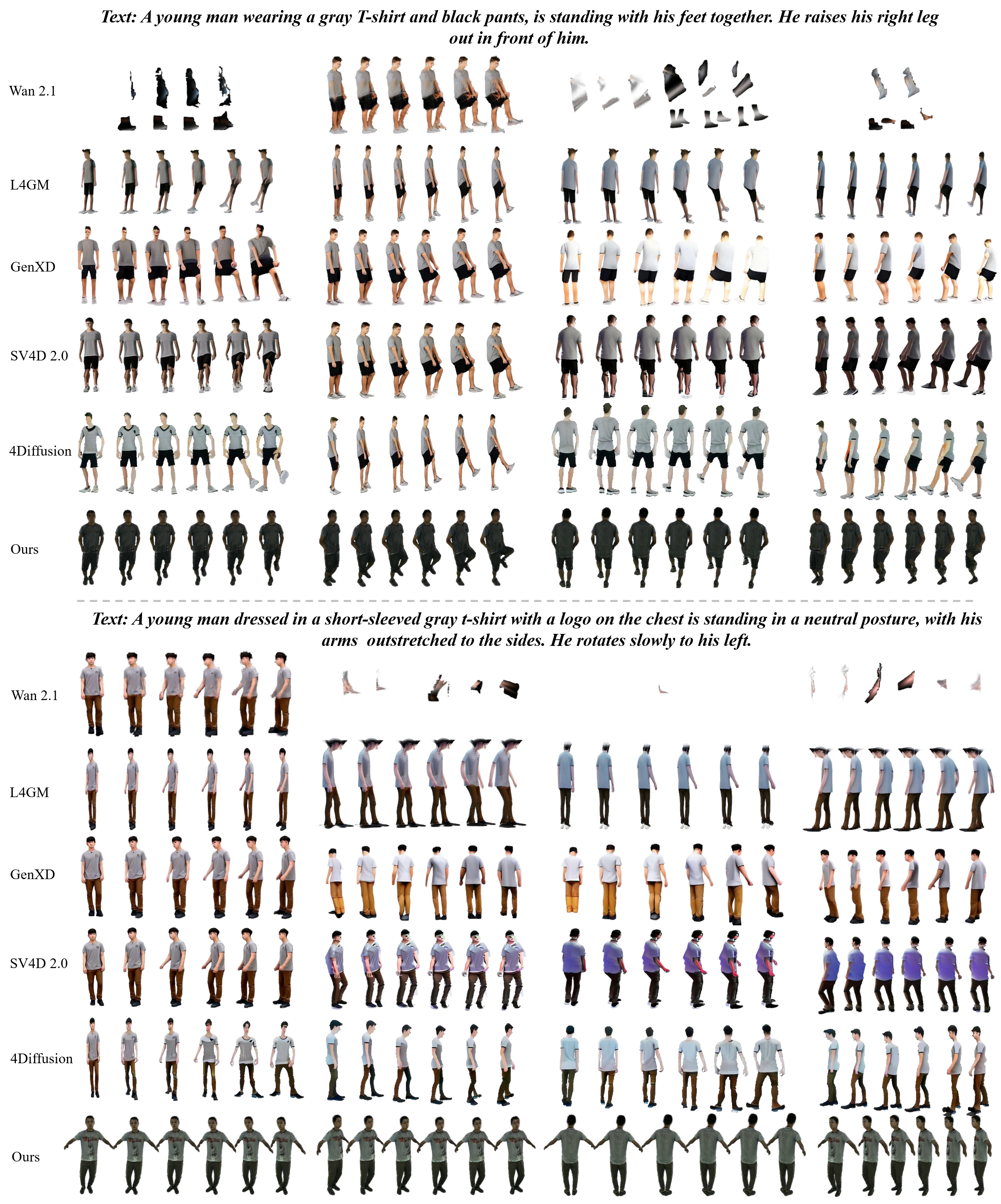}
    \vspace{-1mm}
    \caption{Qualitative comparisons. By directly generating human 4D representations from text prompts, our 4DHumanDiff achieves better consistency along temporal and camera-view dimensions. Please zoom in for better observation.}
    \label{fig:vis_1}
\end{figure*}

\section{Experiments}
\label{sec:experiment}
\subsection{Implementation Details}
In constructing the text-to-4DGS dataset, we use GauHuman~\cite{hu2024gauhuman} to fit each clip for 10k iterations, where the original adaptive density control strategy is replaced by the densification-constrained fitting strategy~\cite{zhang2024gaussiancube}. 
We employ the Adam optimizer~\cite{kingma2014adam} and follow the learning rate setting of GauHuman~\cite{hu2024gauhuman}. 
Each fitting process takes about 15 minutes on a single Nvidia GeForce RTX 2080Ti GPU. 
We set $N_{max}=13824$, $D=24$, and $C=14$.
To standardize the structured 4DGS representation, we compute the mean $\bar{\boldsymbol{\mu}} \in \mathbb{R}^{D \times D \times D \times C}$ and standard deviation $\bar{\boldsymbol{\sigma}} \in \mathbb{R}^{D \times D \times D \times C}$ over the text-to-4DGS dataset, and use them to normalize all training samples.
Text features are extracted using a pretrained CLIP encoder~\cite{radford2021learning}. 
We train 4DHumanDiff with the AdamW optimizer~\cite{loshchilov2017decoupled} on 8 Nvidia GeForce RTX A6000 GPUs for 300k iterations in total, where the 3D U-Net backbone is first trained for 200k iterations. 
The whole training process takes about two weeks. 
We adopt an exponential moving average strategy~\cite{tarvainen2017mean} with a decay rate of 0.9999 to stabilize training. 
The learning rate is decayed from $5\times10^{-5}$ to $1\times10^{-6}$ using cosine annealing~\cite{loshchilov2016sgdr}. 
The batch size is set to 8, and the loss weight $\lambda$ is set to 10. 
For 2D regularization and 4D interpolation, we set the number of rendered views $M$ to 4 and the interpolation factor $W$ to 8.

\subsection{Evaluation Configurations}
We extract 170 text prompts from real-world videos using Gemini 1.5~\cite{team2024gemini} for the evaluation.
To assess the quality of rendered multi-view videos, we adopt several commonly used VBench~\cite{huang2024vbench} metrics, including Image Quality, Subject Consistency, Motion Smoothness, and Overall Consistency.
Image Quality measures visual quality using the no-reference metric MUSIQ~\cite{ke2021musiq}. 
Subject Consistency evaluates appearance stability based on DINO~\cite{caron2021emerging} feature similarity.
Motion Smoothness measures temporal coherence using motion priors from the video interpolation model AMT~\cite{li2023amt}. 
Overall Consistency reflects the semantic alignment between rendered videos and text prompts via ViCLIP~\cite{wang2023internvid}.
In addition, to evaluate multi-view consistency, we measure the appearance stability of rendered multi-view videos at a fixed timestamp using DINO~\cite{caron2021emerging} feature similarity, following a protocol similar to Subject Consistency.

\subsection{Compared with State-of-the-Art Methods}
Existing methods generally follow a two-stage paradigm, where generative models first synthesize monocular or multi-view videos from text prompts, followed by per-scene 4D reconstruction. 
For comparison, we first adopt the recent state-of-the-art text-to-video model Wan 2.1~\cite{wan2025wan} to generate monocular videos from text prompts, and then reconstruct 4D representations for each scene using GauHuman~\cite{hu2024gauhuman}. 
We further compare with the feed-forward 4D reconstruction method L4GM~\cite{ren2024l4gm}, which takes the monocular videos generated by Wan 2.1~\cite{wan2025wan} as input and directly predicts the corresponding 4DGS representations. 
In addition, we consider three methods based on multi-view video diffusion models, including GenXD~\cite{zhao2024genxd}, SV4D2.0~\cite{yao2025sv4d}, and 4Diffusion~\cite{zhang20244diffusion}. 
These methods first generate multi-view videos from the monocular videos synthesized by Wan 2.1~\cite{wan2025wan}, and then perform 4D reconstruction. 
In our experiments, we observe that severe inconsistencies across the generated multi-view videos often lead to failed 4D reconstruction. 
Therefore, for these methods, the comparison is conducted directly on the generated multi-view videos themselves, rather than on videos rendered from reconstructed 4D assets. 
This protocol is more favorable to these compared methods, since it excludes the additional degradation caused by the subsequent 4D reconstruction.

\begin{figure*}[t]
    \centering
    \begin{minipage}[t]{0.49\textwidth}
        \centering
        \includegraphics[width=\linewidth]{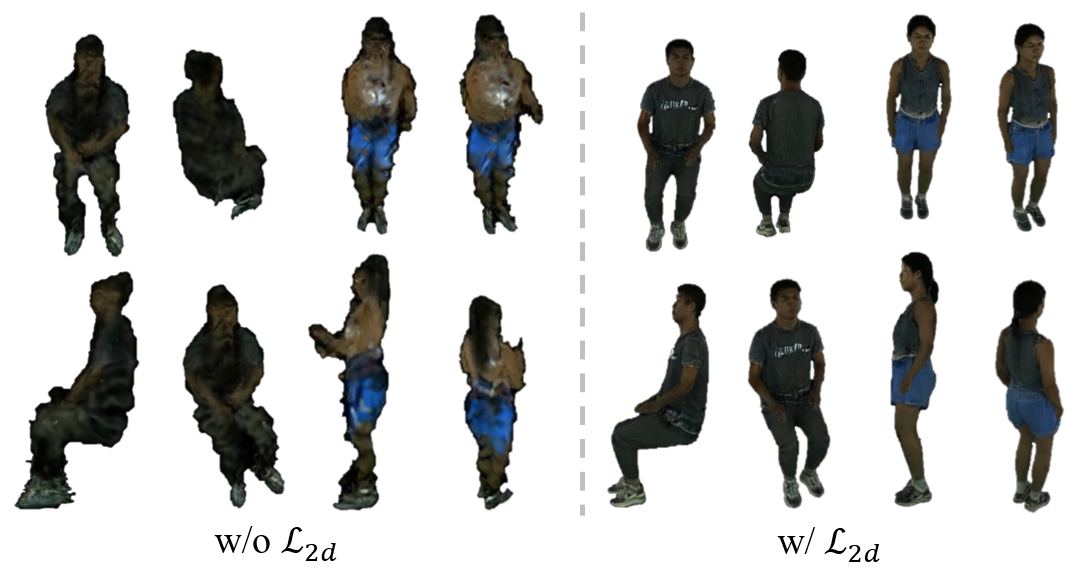}
        \vspace{-3mm}
        \caption{Visual effect of loss terms in 4DHumanDiff  (see \cref{eqn:loss_term}). Without $\mathcal{L}_{2d}$, the model produces results with poor perceptual quality. }
        \label{fig:loss_term}
    \end{minipage}
    \hfill
    \begin{minipage}[t]{0.49\textwidth}
        \centering
        \includegraphics[width=\linewidth]{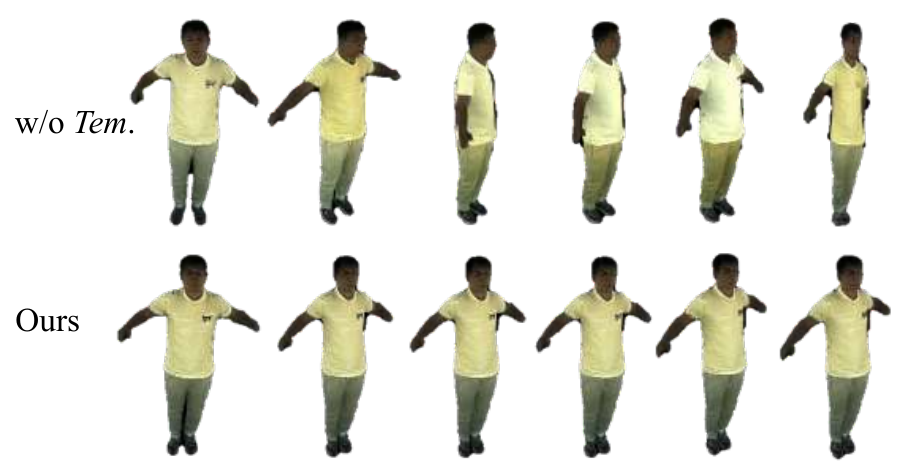}
        \vspace{-3mm}
        \caption{Visual effect of temporal attention. Noticeable appearance changes over time when it is removed (\ie, w/o \textit{Tem.}).}
        \label{fig:effect_of_temporal}
    \end{minipage}
    \vspace{-4mm}
\end{figure*}

\begin{figure}[t!]
    \centering
    \includegraphics[width=0.99\linewidth]{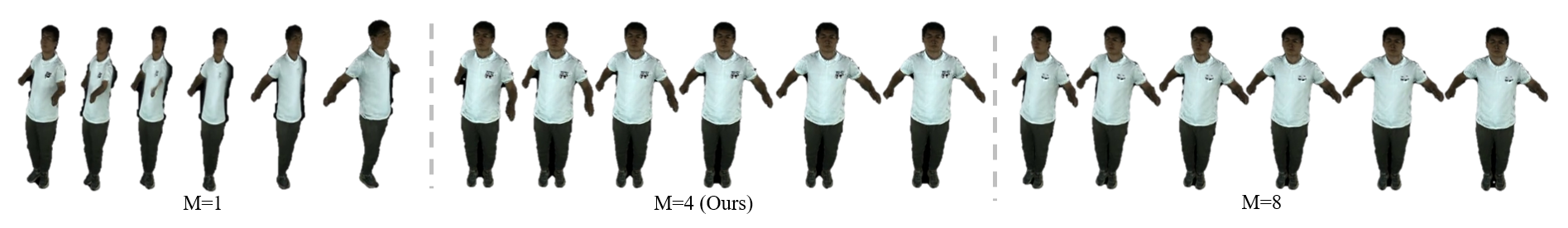}
    \vspace{-2mm}
        \caption{Effect of the view number of rendered videos (\ie, $M$).}
        \label{fig:effect_of_M}
    \vspace{-3mm}
\end{figure}

\noindent\textbf{Quantitative Comparison.}
The quantitative results are summarized in \cref{tab:compared_methods}. 
Overall, 4DHumanDiff achieves strong performance across different evaluation metrics. 
In particular, our method obtains the best scores on Motion Smoothness, Subject Consistency, and Multi-View Consistency, indicating superior consistency in both temporal dynamics and cross-view appearance. 
Although our method does not achieve the highest Image Quality score, we attribute this partly to the tendency of video generation models to produce more vivid colors and sharper local patterns, which are often favored by no-reference quality metrics, as also illustrated in \cref{fig:vis_1}. 
By contrast, our method is designed to prioritize temporally coherent and view-consistent 4D human generation, and therefore shows clearer advantages in consistency-related metrics. 
In addition, video-based pipelines inevitably introduce substantial computational overhead due to video diffusion and per-scene 4D fitting, whereas our method directly generates 4D representations and avoids these expensive intermediate stages, resulting in over 10$\times$ higher efficiency. 
Overall, these results demonstrate that 4DHumanDiff is effective and efficient for generating temporally smooth and view-consistent 4D human assets from text prompts.

\noindent\textbf{Qualitative Comparison.}
The qualitative comparisons are shown in \cref{fig:vis_1}. 
First, 4DHumanDiff generates 4D human content that is well aligned with the input text prompts. 
Second, reconstructing from monocular videos generated by Wan 2.1~\cite{wan2025wan} often fails to recover complete 360-degree human representations due to limited viewpoint coverage. 
Third, the multi-view videos generated by existing multi-view diffusion models exhibit noticeable inconsistencies across large viewpoint changes, mainly because they lack explicit 3D or 4D structural modeling. 
Although some competing methods may appear sharper in individual frontal views, they often fail to maintain stable human structure and appearance under large viewpoint changes. 
In contrast, by directly generating 4D representations from text prompts, our method produces more temporally coherent results and significantly better consistency across viewpoints. 

\noindent\textbf{User Study.}
We further conduct a user study to evaluate  the generated results. 
We randomly select 50 text prompts from the evaluation set and invite 50 participants to perform pairwise comparisons. 
The results of different methods are presented in random order, and the participants are asked to evaluate them from two aspects, \ie, overall quality and multi-view consistency. 
The results show that 4DHumanDiff achieves an average preference rate of 64.8\% on overall quality and 76.4\% on multi-view consistency, both higher than those of competing methods. 
Notably, the advantage of our method is more significant on multi-view consistency, indicating that users prefer our results for their more stable human appearance across different viewpoints, even when some competing methods may produce sharper local details in individual views. 
This again verifies the effectiveness of directly generating structured 4D representations from text prompts.

\begin{figure*}[t!]
    \centering
    \includegraphics[width=0.85\linewidth]{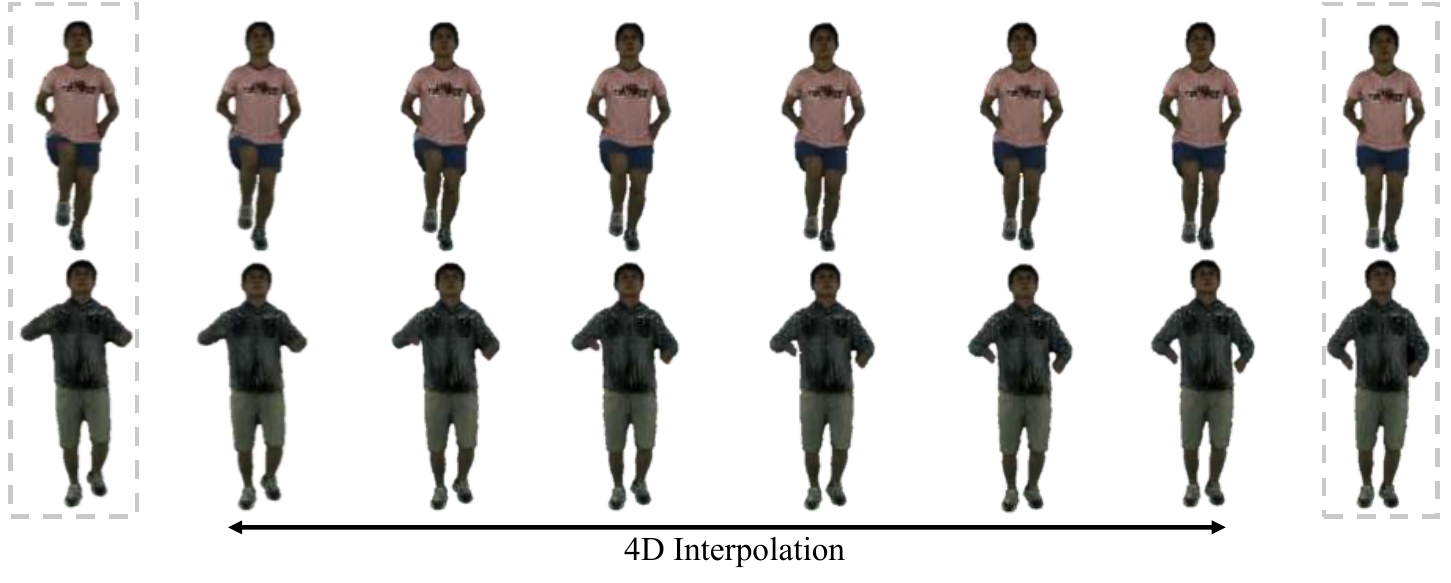}
    \vspace{-4mm}
    \caption{Visual effects of 4D interpolation. Neighboring timestamps without interpolation are indicated by dashed lines. Notably, 4D interpolation increases the frame rate and produces smoother temporal results.}
\label{fig:effect_of_interpolation}
    \vspace{-2mm}
\end{figure*}

\section{Ablation Study}
\label{sec:ablation_studies}

We conduct ablation experiments to validate the effectiveness of the main designs in 4DHumanDiff.
Specifically, we first study the effect of the loss terms used for model optimization. 
We then analyze the role of temporal attention in modeling temporal coherence. 
After that, we evaluate the proposed 4D interpolation strategy for improving motion smoothness. 
Finally, we verify the effectiveness of temporal voxel binding in preserving temporal correspondence within the structured 4DGS representation.

\begin{table*}[t]
\centering
\begin{minipage}[t]{0.49\textwidth}
\centering
\caption{Effect of loss terms. `NaN' implies training collapse.}
\small
\scalebox{0.72}{\begin{tabular}{c c c c c c c}
\toprule
$\mathcal{L}_{4d}$  & $\mathcal{L}_{2d}$  & \tabincell{c}{Image \\ Quality} & \tabincell{c}{Subject \\ Consistency} & \tabincell{c}{Motion \\ Smoothness} & \tabincell{c}{Multi-View \\ Consistency} & \tabincell{c}{Overall \\ Consistency} \\
\midrule
$\times$ & $\checkmark$ & NaN & NaN & NaN & NaN & NaN \\
$\checkmark$ & $\times$ & 0.273 & 0.920 & 0.996 & 0.807 & 0.164 \\
$\checkmark$ & $\checkmark$ & 0.445 & 0.984 & 0.998 & 0.862 & 0.188 \\
\bottomrule
\end{tabular}}
\label{tab:regularization_terms}
\end{minipage}
\hfill
\begin{minipage}[t]{0.49\textwidth}
\centering
\caption{Effect of the view number of rendered videos (\ie, $M$).}
\small
\scalebox{0.78}{\begin{tabular}{c c c c c c}
\toprule
$M$  & \tabincell{c}{Image \\ Quality} & \tabincell{c}{Subject \\ Consistency} & \tabincell{c}{Motion \\ Smoothness} & \tabincell{c}{Multi-View \\ Consistency} & \tabincell{c}{Overall \\ Consistency} \\
\midrule
1 & 0.434 & 0.980 & 0.998 & 0.860 & 0.188 \\
4 & 0.445 & 0.984 & 0.998 & 0.862 & 0.188 \\
8 & 0.445 & 0.983 & 0.998 & 0.862 & 0.188 \\
\bottomrule
\end{tabular}}
\label{tab:effect_of_M}
\end{minipage}
\vspace{-3mm}
\end{table*}

\begin{table*}[t]
\centering
\begin{minipage}[t]{0.49\textwidth}
\centering
\caption{Effect of excluding background patches in $\mathcal{L}_{vgg}$.}
\small
\scalebox{0.7}{\begin{tabular}{c c c c c c}
\toprule
Methods  & \tabincell{c}{Image \\ Quality} & \tabincell{c}{Subject \\ Consistency} & \tabincell{c}{Motion \\ Smoothness} & \tabincell{c}{Multi-View \\ Consistency} & \tabincell{c}{Overall \\ Consistency} \\
\midrule
w/o \textit{Exc.} & 0.430 & 0.975 & 0.997 & 0.862 & 0.186 \\
Ours & 0.445 & 0.984 & 0.998 & 0.862 & 0.188 \\
\bottomrule
\end{tabular}}
\label{tab:patch_vgg}
\end{minipage}
\hfill
\begin{minipage}[t]{0.49\textwidth}
\centering
\caption{Effect of temporal attention layers.}
\small
\scalebox{0.7}{\begin{tabular}{c c c c c c}
\toprule
Methods & \tabincell{c}{Image \\ Quality} & \tabincell{c}{Subject \\ Consistency} & \tabincell{c}{Motion \\ Smoothness} & \tabincell{c}{Multi-View \\ Consistency} & \tabincell{c}{Overall \\ Consistency} \\
\midrule
w/o \textit{Tem.} & 0.429 & 0.961 & 0.996 & 0.862 & 0.183 \\
Ours & 0.445 & 0.984 & 0.998 & 0.862 & 0.188 \\
\bottomrule
\end{tabular}}
\label{tab:temporal_attention}
\end{minipage}
\end{table*}

\begin{table*}[t]
\centering
\begin{minipage}[t]{0.49\textwidth}
\centering
\caption{Effect of 3D U-Net pretraining.}
\small
\scalebox{0.7}{\begin{tabular}{c c c c c c}
\toprule
Methods & \tabincell{c}{Image \\ Quality} & \tabincell{c}{Subject \\ Consistency} & \tabincell{c}{Motion \\ Smoothness} & \tabincell{c}{Multi-View \\ Consistency} & \tabincell{c}{Overall \\ Consistency} \\
\midrule
w/o \textit{Pre.} & 0.445 & 0.984 & 0.998 & 0.860 & 0.185 \\
Ours & 0.445 & 0.984 & 0.998 & 0.862 & 0.188 \\
\bottomrule
\end{tabular}}
\label{tab:two_stage}
\end{minipage}
\hfill
\begin{minipage}[t]{0.49\textwidth}
\centering
\caption{Effect of temporal voxel binding.}
\small
\scalebox{0.7}{\begin{tabular}{c c c c c c}
\toprule
Methods & \tabincell{c}{Image \\ Quality} & \tabincell{c}{Subject \\ Consistency} & \tabincell{c}{Motion \\ Smoothness} & \tabincell{c}{Multi-View \\ Consistency} & \tabincell{c}{Overall \\ Consistency} \\
\midrule
w/o Binding & 0.443 & 0.973 & 0.996 & 0.861 & 0.186  \\
Ours & 0.445 & 0.984 & 0.998 & 0.862 & 0.188\\
\bottomrule
\end{tabular}}
\label{tab:temporal_voxelization}
\end{minipage}
\end{table*}

\subsection{Effect of Loss Terms}
As shown in \cref{eqn:loss_term}, 4DHumanDiff is optimized with both 4D and 2D supervision terms, namely $\mathcal{L}_{4d}$ and $\mathcal{L}_{2d}$.
The quantitative effects of these two loss terms are summarized in \cref{tab:regularization_terms}. 
Without $\mathcal{L}_{4d}$, the model becomes unstable during optimization and fails to converge, resulting in invalid outputs. 
Without $\mathcal{L}_{2d}$, the model can still converge stably, but the perceptual quality of the rendered results is noticeably degraded. 
Combining the two terms yields the best performance across all metrics. 
The qualitative results in \cref{fig:loss_term} further show that $\mathcal{L}_{2d}$ significantly improves the visual quality of the rendered images.

To compute $\mathcal{L}_{2d}$, we render videos from $M$ viewpoints. 
The quantitative effect of different choices of $M$ is reported in \cref{tab:effect_of_M}.
In general, increasing $M$ leads to better performance, since it implicitly imposes stronger constraints on multi-view consistency. 
As shown in \cref{fig:effect_of_M}, a smaller view number (\ie, $M=1$) tends to produce flattened body shapes under twisting motions, while increasing $M$ effectively alleviates this issue. 
We set $M$ to 4 by default.

Besides, we exclude patches that do not contain human regions when computing the perceptual features, so as to reduce the interference from background content. 
As shown in \cref{tab:patch_vgg}, restricting the perceptual loss to human-related regions leads to better performance, indicating that this design provides more targeted supervision for modeling fine-grained human appearance and motion.

\subsection{Effect of Temporal Attention}
\cref{tab:temporal_attention} shows the effect of temporal attention layers. 
Without temporal attention (\ie, w/o \textit{Tem.}), the model yields noticeably lower Subject Consistency and Overall Consistency, indicating weaker temporal coherence. 
The qualitative results in \cref{fig:effect_of_temporal} further show that removing temporal attention leads to appearance variations over time.

We further study the effect of the two-stage training strategy for the 4D U-Net, where the 3D U-Net backbone is first trained and the temporal attention layers are then jointly optimized. 
As shown in \cref{tab:two_stage}, removing the pre-training stage for the 3D U-Net (\ie, w/o \textit{Pre.}) leads to a drop in Overall Consistency, indicating weaker semantic alignment between the rendered videos and the input text prompts.

\subsection{Effect of 4D Interpolation}
The effect of the proposed training-free 4D interpolation is illustrated in \cref{fig:effect_of_interpolation}. 
The neighboring timestamps without interpolation are marked by dashed lines. 
By inserting intermediate 4DGS states between adjacent timestamps, this strategy effectively increases the frame rate and produces temporally smoother results without additional training. 
Moreover, the interpolation factor $W$ can be flexibly adjusted according to the desired temporal resolution.

\subsection{Effect of Temporal Voxel Binding}

To validate the effectiveness of our temporal voxel binding strategy, we compare it with an alternative design that independently voxelizes the Gaussians at each timestamp (\ie, w/o Binding). 
In the baseline setting, each temporal slice is organized into the voxel grid separately, without explicitly preserving temporal correspondence across timestamps. 
By contrast, our method performs voxelization only once at the first timestamp and assigns Gaussians at the remaining timestamps to the same grid positions according to the temporal binding relationship.

The quantitative results are shown in \cref{tab:temporal_voxelization}. 
Compared with independent voxelization, our method consistently improves all evaluation metrics, with more noticeable gains in Subject Consistency, Motion Smoothness, and Overall Consistency. 
These results indicate that explicitly preserving temporal correspondence in the structured 4DGS representation is beneficial for stable dynamic human generation. 
In particular, by enforcing all timestamps to share the same voxel layout, our method better maintains the structural identity of the human body over time, which leads to more coherent motion and more stable appearance across views.

\section{Limitations}

This work mainly focuses on direct text-to-4DGS generation of temporally coherent and view-consistent 360-degree dynamic humans. 
While the proposed framework already achieves strong overall 4D consistency, incorporating stronger fine-grained local detail enhancement, especially for challenging articulated regions, remains an important direction for future work.
In particular, the current framework does not explicitly introduce dedicated local-detail refinement modules or fine-scale geometric priors. 
Exploring how to further improve local human geometry and appearance while preserving global temporal and cross-view consistency would be a valuable next step.

\section{Conclusions}
\label{sec:conclusion}
In this paper, we presented 4DHumanDiff, a diffusion framework that directly generates 360-degree dynamic humans represented by 4D Gaussian Splatting (4DGS) from text prompts. 
Different from existing video-first and reconstruction-later pipelines, our method formulates dynamic human generation directly in the structured 4D representation space, which is better matched to the spatial organization and temporal regularity of human motion.
Built on a 3D U-Net backbone with temporal self-attention, 4DHumanDiff enables motion-aware generation in a unified framework. 
To support this task, we further constructed a large-scale text-to-4DGS dataset with 60,000 high-quality pairs for training from scratch.
We also introduced a 2D regularization term and a training-free 4D interpolation strategy to improve rendering quality and motion smoothness. 
Extensive experiments show that 4DHumanDiff achieves more consistent results across time and viewpoints while reducing the overall generation time by more than 10$\times$ compared with existing two-stage pipelines.

\section{Acknowledgements}
This work was supported by the National Natural Science Foundation of China under Grant No.~62371164 and the National Key RD Program of China under Grant No.~2022YFA1004100.

\clearpage




\end{document}